\documentclass{article}

\usepackage[final]{neurips_2025}

\usepackage[utf8]{inputenc} 
\usepackage[T1]{fontenc}    
\usepackage{hyperref}       
\usepackage{url}            
\usepackage{booktabs}       
\usepackage{amsfonts}       
\usepackage{nicefrac}       
\usepackage{microtype}      
\usepackage{xcolor}         
\usepackage{microtype}
\usepackage{hyperref}
\usepackage{url}
\usepackage{booktabs}
\usepackage{graphicx} 
\usepackage{natbib}
\usepackage[]{xcolor}
\usepackage{tikz}
\usepackage{pgfplots}
\usepackage{booktabs}
\usepackage{amsmath}
\usepackage{amssymb}
\usepackage{caption}
\usepackage{stfloats}
\usepackage{hyperref}
\usepackage{soul}
\usepackage{lineno}
\usepackage{wrapfig,lipsum,booktabs}

\definecolor{darkblue}{rgb}{0, 0, 0.5}
\hypersetup{colorlinks=true, citecolor=darkblue, linkcolor=darkblue, urlcolor=darkblue}

\title{Neural Networks for Learnable and Scalable Influence Estimation of Instruction Fine-Tuning Data}

\author{Ishika Agarwal \& Dilek Hakkani-Tür\\
Department of Computer Science\\
UIUC\\
Urbana-Champaign, IL 61801, USA \\
\texttt{\{ishikaa2, dilek\}@illinois.edu} \\
}

\definecolor{myorange}{RGB}{255, 153, 0}
\definecolor{mygreen}{RGB}{78, 167, 46}
\newcommand{\hlc}[2][yellow]{{%
    \colorlet{foo}{#1}%
    \sethlcolor{foo}\hl{#2}}%
}
\newcommand{\sysn}{NN-CIFT}

\begin{document}

\maketitle

\begin{abstract}
Influence functions provide crucial insights into model training, but existing methods suffer from large computational costs and limited generalization. Particularly, recent works have proposed various metrics and algorithms to calculate the influence of data using language models, which do not scale well with large models and datasets. This is because of the expensive forward and backward passes required for computation, substantial memory requirements to store large models, and poor generalization of influence estimates to new data. In this paper, we explore the use of small neural networks -- which we refer to as the InfluenceNetwork -- to estimate influence values, achieving up to 99\% cost reduction. Our evaluation demonstrates that influence values can be estimated with models just 0.0007\% the size of full language models (we average across 1.5B-22B versions). We apply our algorithm of estimating influence values (called \textbf{\sysn{}}: \textbf{N}eural \textbf{N}etworks for effi\textbf{C}ient \textbf{I}nstruction \textbf{F}ine-\textbf{T}uning) to the downstream task of subset selection for general instruction fine-tuning. In our study, we include four state-of-the-art influence functions and show no compromise in performance, despite large speedups, between \sysn{} and the original influence functions. We provide an in-depth hyperparameter analyses of \sysn{}. The code for our method can be found here: \href{https://github.com/agarwalishika/NN-CIFT}{https://github.com/agarwalishika/NN-CIFT}.
\end{abstract}

\begin{figure*}
    \centering
    \includegraphics[width=\linewidth]{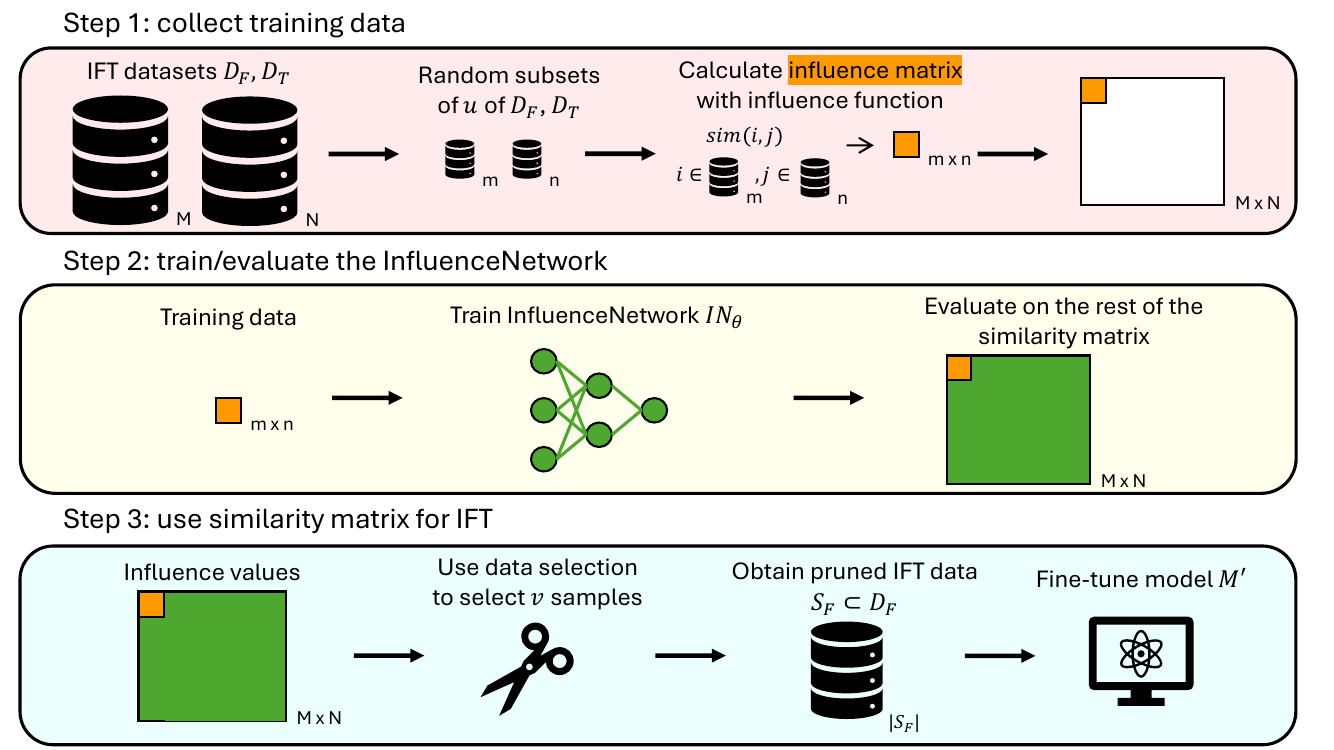}
    \caption{Overview of \sysn{}. The first step consists of using established influence functions to \hlc[myorange]{collect data} for training the InfluenceNetwork. Next, the data from Step (1) is used to \textbf{train the InfluenceNetwork} and, subsequently, \hlc[mygreen]{estimate the influence values} for the rest of the data. Finally, the data selection algorithm corresponding to the original influence function is used to \textbf{select a subset of IFT data} to fine-tune a model on.}
    \label{fig: nncift}
\end{figure*}

\section{Introduction}

The strong instruction-following abilities of large language models (LLMs) can be attributed to instruction fine-tuning (IFT) \citep{zhang2024instructiontuninglargelanguage}. IFT builds on top of current language modeling capabilities and strengthens the instruction following abilities of models. Recent works have taken \textbf{data efficient} approaches for IFT. The goal is to select a small subset of samples on which to fine-tune a model \cite{delift, craig, deftucs, less, smart, tsds} that emulates the full dataset.

Data efficient pipelines typically consist of two stages: (1) \textit{data valuation}: designing functions to estimate the influence of data points, and (2) \textit{data selection}: using influence estimates to choose a balanced set of influential data. Usually, data selection is cheaper than valuation -- for instance, DELIFT (SE)\footnote{Short for "Sentence Embedding".} \citep{delift} computes the similarity of sentence embeddings between pairs of data (expensive) for valuation and selects representative data using a submodular function (cheap).

Formally, influence functions estimate the value of data. For instance, brute force influence functions use leave-one-out (LOO) training to measure impact by omitting each data point and evaluating performance \cite{loo_1982}. More recent influence functions use LLMs to estimate influence. Table \ref{table: costs} outlines the expenses of state-of-the-art (SOTA) influence functions, which comes from the large amount of forward and backward passes through highly parameterized models.

\begin{wraptable}{r}{8cm}
    \resizebox{0.58\columnwidth}{!}{
    \begin{tabular}{lcc}
    \toprule
    Method & Cost & Size \\
    \midrule
    Pairwise \\
    \midrule
        \hspace{1mm} DELIFT \citep{delift} & $\mathcal{O}(MN) \cdot F$ & 7-8B \\
        \hspace{1mm} DELIFT (SE) \citep{delift} & $\mathcal{O}(MN) \cdot F$ & 355M \\
        \hspace{1mm} LESS \citep{less} & $\mathcal{O}(M+N) \cdot B$ & 7-8B \\
        \hspace{1mm} \sysn{} (ours) & $\mathcal{O}(MN) \cdot F$ & 205K \\
    \midrule
    Pointwise \\
    \midrule
        \hspace{1mm} SelectIT \citep{selectit} & $\mathcal{O}(M) \cdot F$ & 7-8B \\
    
        \hspace{1mm} \sysn{} (ours) & $\mathcal{O}(M) \cdot F$ & 205K \\
    \bottomrule
    \end{tabular}
    }
    \caption{Approximating the computational complexity of data valuation in terms of the cost of forward passes ($F$) or backward passes ($B$) through a model. $M = |\mathcal{D_F}|$ and $N = |\mathcal{D_T}|$, a fine-tuning and target dataset respectively, we use for subset selection. See Appendix \ref{app: pairwise influence functions} for more details. Size denotes the number of parameters of the corresponding model. Note: larger models have a higher $F$ and $B$.}
    \label{table: costs}
\end{wraptable}

In this paper, we introduce \textbf{\sysn{}}: \textbf{N}eural \textbf{N}etworks for effi\textbf{C}ient \textbf{I}nstruction \textbf{F}ine-\textbf{T}uning and explore how to train influence functions efficiently. We improve efficiency by using compact neural networks -- which we coin as the InfluenceNetwork -- that are 0.0077\% the size of LLMs, to estimate influence. Figure \ref{fig: nncift} outlines our methodology with a pairwise influence function (more details about pairwise influence functions in Appendix \ref{app: pairwise influence functions}).

As depicted, \sysn{} is a three-step algorithm. The neural network must be trained to estimate influence values effectively. Hence, we first use the influence function (with LLMs) to output influence values for \hlc[myorange]{a very small subset of data}. This becomes our training data for the InfluenceNetwork. We find that a small neural network can sufficiently learn to estimate influence with very few data (covered in Section \ref{sec: learning_influence_estimation}).

Second, we train the InfluenceNetwork, and use it to estimate the influence values \hlc[mygreen]{for the rest of the data points}. Finally, we apply a data selection algorithm on the influence values. This helps to obtain a small subset of IFT data to enhance language models. After fine-tuning language models on the chosen subsets, we find that \sysn{} achieves comparable performance to the original influence functions (covered in Section \ref{sec: subset_selection_evaluation}). 

Our contributions and findings are listed as follows. \sysn{}:
\begin{enumerate}
    \item \textbf{alleviates the cost of using expensive LLMs during data valuation} by using smaller and cheaper neural networks, without affecting the performance on downstream tasks (Tables \ref{table: llama v=0.3}-\ref{table: mistral v=0.3});
    \item \textbf{achieves competitive performance to previous data valuation methods, despite using only 0.25\%-5\% of the data.} The average mean square error in influence values between \sysn{} and the original influence functions is merely 0.067 (Figure \ref{fig: influence_net});
    \item is shown to be effective for new data points, \textbf{circumventing the need to retrain an influence function for new data} -- previous works incur this cost (Figure \ref{fig: influence_net}).
    \item \textbf{reduces costs by 77-99\% time} during data valuation (Table \ref{table: actual_costs}).
\end{enumerate}

Section \ref{sec: related works} outlines the current state of research in data valuation and data selection. Section \ref{sec: influence functions} explains the problem setting. Section \ref{sec: learning_influence_estimation} presents the main methodology for \sysn{} and motivating results. Finally, Section \ref{sec: subset_selection_evaluation} reports results on the downstream task of subset selection after the data valuation stage. In our evaluation, we find that using a small LLM with the original influence functions results in degraded performance. Our hyperparameter studies are in Appendix \ref{subsec: in_size}, Figure \ref{fig: influence_network_size} and Appendix \ref{subsec: hp_u_v}, Figure \ref{fig: hyperparam_study}. We also show language model performance with smaller subsets of selected fine-tuning data in Appendix \ref{app: smaller_subset_eval}. Lastly, the SOTA influence functions are detailed in Appendix \ref{app: influence_functions}.

\section{Related Works}
\label{sec: related works}
\subsection{Data Valuation}
\label{sec: data valuation}
\citet{wei2023largerlanguagemodelsincontext} hint that different models extract different information from the same data. Hence, effective fine-tuning requires datasets to be specific to each model. Not all data points affect the model equally - models learn more from certain data points than others. Therefore, data valuation methods prune out such low-influence data for efficient fine-tuning \citep{less, delift}. Current research is divided into model-independent and model-dependent valuation metrics. 

Model-independent methods, such as distance or clustering-based methods \citep{deftucs, tsds, smart} are faster and less computationally expensive. Distance-based methods assign more "influence" to data points that are further from each other, optimizing for a diverse subset. Clustering-based methods assign more "influence" to data points that are representative (i.e., the centroids of clusters). 

On the other hand, model-dependent methods -- such as inference-based and gradient-based -- are more resource intensive. Inference-based methods \citep{selectit, delift} use model inference signals (e.g., token distributions) to evaluate the performance or confidence of models, and valuate data based on how performative/confident they are. Gradient based methods \citep{less, craig, gradmatch, kohliang}, on the other hand, can assign higher influence to data points with (1) higher magnitudes of gradients, or (2) gradients that match domain-specific data (for domain-specific fine-tuning, for example).

While they are expensive to calculate, when paired with data selection algorithms, model-dependent data valuation metrics can be used to select subsets of data that are specific to a model's capabilities. Model-dependent data valuation metrics help to select data that will maximize a certain objective for each model, rendering fine-tuning more effective. 

\subsection{Data Selection}
Data selection aims to prune redundant and noisy data samples from large datasets to produce a small, information-rich subset \citep{delift, less}. This subset should be representative of the larger dataset while performing comparably, if not better, than using the full dataset. Data selection methods usually have objectives for selecting data: (1) instruction tuning \citep{selectit}, (2) task-specific fine-tuning \citep{tsds}, (3) continual learning \citep{delift}, (4) preference alignment \citep{deita}, etc. While certain objectives are subsets of others (e.g. (2) is subset of (1)), the data selected for each purpose may not necessarily overlap. For instance, (1) requires data that is representative of a particular dataset, whereas (2) focuses on samples that reflect specific tasks like math reasoning, question answering, or summarization. Similarly, (3)'s samples are specifically chosen to introduce new information to a model without overriding or repeating previously learned information.

\section{Problem Formulation}
\label{sec: influence functions}
Given a model $\mathcal{M}$ and fine-tuning data $\mathcal{D_F}$, the goal is to select a small subset $\mathcal{S_F} \subset \mathcal{D_F}$ that maximizes the performance of $\mathcal{M}$ after fine-tuning $\mathcal{M}$ on $\mathcal{S_F}$. $\mathcal{S_F}$ is the optimal subset if it can be used to train a model that is comparable to a model trained on $\mathcal{D_F}$. However, more recent works jointly optimize other objectives during subset selection. Examples of objectives include not only representation, but also task-specific refinement and continual learning. For such joint optimization, the subset $\mathcal{S_F}$ is aligned with another target domain dataset $\mathcal{D_T}$. The choice of $\mathcal{D_T}$ can guide the subset selection towards various objectives.
For example, if the objective is representation or task-specific refinement, $\mathcal{S_F}$ will contain points from $\mathcal{D_F}$ that are similar to $\mathcal{D_T}$ \citep{tsds, less, deftucs}. Alternatively, if the objective is continual learning, $\mathcal{S_F}$ will contain points from $\mathcal{D_F}$ that would allow the model $\mathcal{M}$ to learn new information that is present in $\mathcal{D_T}$ \cite{delift, gcr}.

As mentioned before, computing influence functions can be a very expensive process. There are two kinds of influence functions: pairwise and pointwise -- both require forward/backward passes through language models, but the costs slightly differ. Pairwise influence functions compute the influence between every pair of points in a dataset. We study three SOTA pairwise functions, whose formulations are details in Appendix \ref{app: pairwise influence functions}. This paper also studies one pointwise influence functions that simply compute the influence of each data point individually, formally outlined in Appendix \ref{app: pointwise_influence_functions}. While pointwise influence functions are more efficient than pairwise, they are not as performant during subset selection \cite{less, delift}.


\subsection{Our motivation}
Overall, our aim is to reduce the total number of forward or back propagations through models with millions and billions of parameters by replacing a large portion with forward propagations through small neural networks with (merely) hundreds of thousands of parameters. Pairwise influence functions calculate the similarity between two data points (denoted as $sim(i, j)$). Because influence values are usually not learned, they need to be recomputed for any data beyond the training data. In other words, as data is constantly being collected, influence values for new data must be recomputed. However, \sysn{} is learned. Hence, \textit{our method does not require any extra computation to estimate influence values}, unlike previous work.

\begin{figure*}[]
    \centering
    \includegraphics[width=\linewidth]{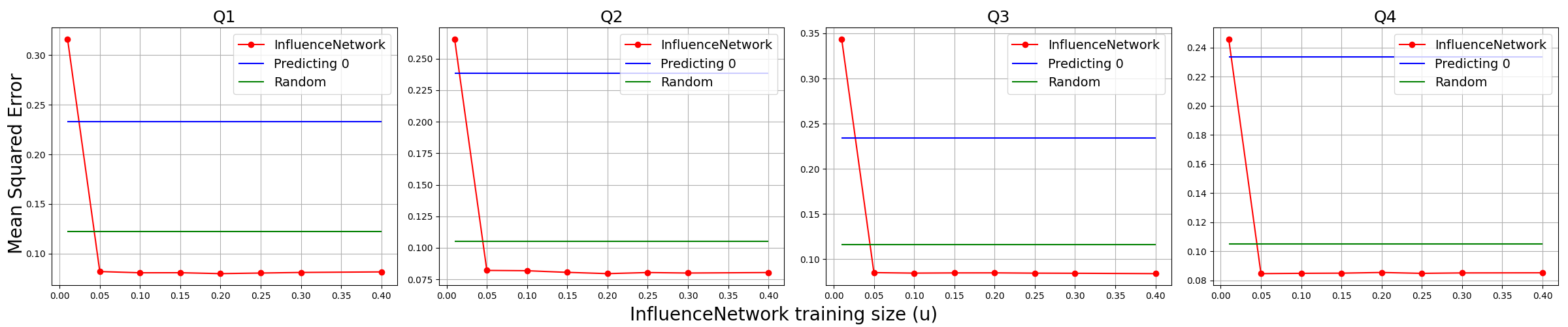}
    \caption{MSE versus InfluenceNetwork training data size (u) plotted for 8 different training sizes, broken down by the quadrant. These results are for learning DELIFT influence values. Error rates on each quadrant correspond to losses across different sets: Q1 for training, Q2/Q3 for validation, and Q4 for testing. As shown, the InfluenceNetwork achieves MSE of merely 0.05\% starting from $u=0.05$ and always outperforms the baselines.}
    \label{fig: influence_net}
\end{figure*}

\section{Learning Influence Estimation}
\label{sec: learning_influence_estimation}
This section describes in detail Steps 1 and 2 in Figure \ref{fig: nncift}. It outlines the structure and initial experimentation of the InfluenceNetwork.

\subsection{Defining the InfluenceNetwork.} 
For estimating the influence values of data samples, we call our neural network the \textit{InfluenceNetwork}. It is a 2-layer neural network with a hidden size of 100 neurons, and an output size of 1 neuron. For activation, we use ReLU in between the layers. The function $IN_\theta$ represents the neural network with parameters $\theta$. As input, $IN_\theta$ takes two data points $i$ and $j$ and outputs the estimated influence of $i$ on $j$. Specifically, embeddings for $i$ and $j$ are computed (denoted as \texttt{emb}() below) using the BAAI General Embedding model (\texttt{bge-large-en-v1.5}, in particular) \citep{bge_embedding} and are concatenated:

\begin{align}
0 \leq IN_\theta(i, j)&\leq 1, \label{equation: influence_net_formulation} \\
0 \leq \theta(\texttt{concat}(\texttt{emb}(i), \texttt{emb}(j)))&\leq 1, \\
\forall (i, j) \in \mathcal{D_F} \times \mathcal{D_T}
\end{align}

\noindent The \texttt{bge-large-en-v1.5} model generates embeddings of size 1,024, which means the input has a total length of 2,048. Hence, the InfluenceNetwork has exactly 204,900 parameters. For training, we use 20 epochs and a learning rate $\eta = 0.0001$.

We note that \sysn{} relies on an embedding model that is often larger than the actual size of the InfluenceNetwork. However, we choose not to include it in the cost for a few reasons: (1) our method does not rely on the underlying embedding model, (2) NLP pipelines generally use embedding models to store, retrieve, cluster, and/or visualize data and hence, is an offline cost.

\subsection{Training the InfluenceNetwork.}

Below is an illustration of the quadratic similarity matrix that is computed during the data valuation stages. Previous influence compute the entire matrix for data valuation -- we only use Q1.

Using the predefined influence functions in Appendix \ref{app: influence_functions}, a small fraction of influence values are computed -- we call this fraction $u$. We use $u$\% of data from $\mathcal{D_F}$ and $u$\% of data from $\mathcal{D_T}$ to compute the training set for the InfluenceNetwork. As mentioned above, this training set is represented by Q1 in the illustration.

The quadrants Q1 to Q4 represent the subset of influence values between a combination of in-distribution (ID) data and out-of-distribution (OOD) data. ID and OOD data is determined by whether the InfluenceNetwork was trained on the data (ID) or not (OOD):
\begin{wraptable}{r}{6cm}
\vspace{-10pt} 
\centering
    \scalebox{0.8}{ 
    \begin{tikzpicture}
    \draw[step=1cm,color=gray] (-1,-1) grid (1,1);
    \node at (-0.5,+0.5) {Q1};
    \node at (+0.5,+0.5) {Q2};
    \node at (-0.5,-0.5) {Q3};
    \node at (+0.5,-0.5) {Q4};
    \draw[-latex] (-1.5,-1.5)--(-1.5,1.5) node[pos=1.1]{$\mathcal{D_F}$};
    \draw[-latex] (-1.5,-1.5)--(1.5,-1.5) node[pos=1.1]{$\mathcal{D_T}$};
    \end{tikzpicture}
    }
    \vspace{-40pt} 
\end{wraptable}
\begin{itemize}
    \item Q1: Fully ID data from $\mathcal{D_F}$ and $\mathcal{D_T}$
    \item Q2: ID data from $\mathcal{D_F}$ and OOD data from $\mathcal{D_T}$ 
    \item Q3: OOD data from $\mathcal{D_F}$ and ID data from $\mathcal{D_T}$
    \item Q4: Fully OOD data from $\mathcal{D_F}$ and $\mathcal{D_T}$
\end{itemize}

\subsection{Evaluating the InfluenceNetwork.}
To ensure our InfluenceNetwork is able to output influence values correctly, we compute the average mean squared error (MSE) between the ground truth influence values (from Appendix \ref{app: influence_functions}) and the predicted influence values: 
\begin{align*}
\frac{1}{|\mathcal{D_F} \times \mathcal{D_T}|} \sum_{(i, j) \in \mathcal{D_F} \times \mathcal{D_T}} (IF_\theta(i, j) - \text{sim}(i, j)) ^ 2
\end{align*}

\noindent We separate the evaluation between the four quadrants of data to study the performance with ID and OOD data.

To train the InfluenceNetwork, we use DELIFT's influence values on the MixInstruct dataset \citep{mixinstruct} to train our InfluenceNetwork (more dataset details in Section \ref{sec: subset_selection_evaluation}). We report the results from \textit{InfluenceNetwork} and two other baselines: (1) \textit{Random}ly generating a number between 0 and 1, and (2) only \textit{Predicting 0} influence. These results can be found in Figure \ref{fig: influence_net}.

The InfluenceNetwork is able to predict influence values with low error rates. After just $u=0.05$, it is consistently better than random influence values and predicting only 0. The average MSE between the InfluenceNetwork's influence scores and DELIFT's influence scores is 0.072, 0.072, 0.062, 0.063 for Q1 to Q4, respectively (averaging to 0.067). Furthermore, \textbf{the error rate stays consistent across all four quadrants, showing that \sysn{} does not need to be retrained to estimate the influence of new data points} that are collected after the training data. One thing to note is that \textit{although $u=0.05$, with pairwise influence functions, we end up using only 0.25\% of the data} to train the InfluenceNetwork because we use 5\% of $\mathcal{D_F}$ and 5\% of $\mathcal{D_T}$. 

Finally, we report the robustness  of the InfluenceNetwork. In Table \ref{table: influence_network_variance} reports the variance of the InfluenceNetwork after five runs. We assume the original influence values are given and static, and therefore, do not measure the variance of the individual influence functions. Table \ref{table: influence_network_variance} shows low variance across each quadrant and each $u$ in the InfluenceNetwork scores. Table \ref{table: embedding_model} shows that the InfluenceNetwork is invariant to the choice of the embedding model. As the MSE between the predicted influence values and ground truth influence values remains small, we posit the downstream performance of NN-CIFT will be maintained no matter the choice of embedding model.

\begin{table}[]
\centering\scriptsize
\begin{tabular}{c|cccc}
\toprule
u    & Q1 std  & Q2 std & Q3 std & Q4 std \\
\midrule
0.01 & 0.0050  & 0.0047 & 0.0018 & 0.0118 \\
0.05 & 0.0090  & 0.0091 & 0.0032 & 0.0032 \\
0.10 & 0.0062  & 0.0058 & 0.0032 & 0.0032 \\
0.15 & 0.0062  & 0.0056 & 0.0037 & 0.0037 \\
0.20 & 0.0102  & 0.0096 & 0.0020 & 0.0020 \\
0.25 & 0.0102  & 0.0096 & 0.0020 & 0.0020 \\
0.30 & 0.0118  & 0.0112 & 0.0020 & 0.0020 \\
0.40 & 0.0067  & 0.0053 & 0.0018 & 0.0018 \\
\bottomrule 
\end{tabular}
\caption{Variance of the InfluenceNetwork for varying $u$'s for DELIFT values.}
\label{table: influence_network_variance}
\end{table}

\begin{table}[]
\centering\scriptsize
\begin{tabular}{c|cccc}
\toprule
Embedding Model & Q1  & Q2  & Q3  & Q4 \\
\midrule
BAAE/bge-large-en-v1.5 & 0.051 & 0.084 & 0.074 & 0.084 \\
Qwen/Qwen3-Embedding-0.6B & 0.076 & 0.087 & 0.087 & 0.089 \\
intfloat/e5-mistral-7b-instruct & 0.026 & 0.083 & 0.084 & 0.083 \\
Snowflake/snowflake-arctic-embed-l-v2.0 & 0.077 & 0.087 & 0.087 & 0.088 \\
\bottomrule 
\end{tabular}
\caption{Varying embedding models and their corresponding MSE values (averaged across 5 runs) between estimated influence values and ground truth influence values, with 5\% selected data. These results are for learning DELIFT influence values. As shown, NN-CIFT is invariant to the embedding model selected, and is able to effectively estimate the influence values.}
\label{table: embedding_model}
\end{table}

\begin{table*}[th]
\centering\scriptsize
\makebox[\linewidth][c]{%
\begin{tabular}{lcccccccccccccc}
\toprule
Dataset                 & \multicolumn{6}{c}{MixInstruct}                                                                                                         & \multicolumn{6}{c}{Alpaca}   & \multicolumn{2}{c}{MMLU}                                                                                                           \\ \cmidrule(lr){2-7} \cmidrule(lr){8-13} \cmidrule{14-15}
Method                & \multicolumn{3}{c}{ICL} & \multicolumn{3}{c}{QLoRA} & \multicolumn{3}{c}{ICL} & \multicolumn{3}{c}{QLoRA} & \multicolumn{1}{c}{ICL} & \multicolumn{1}{c}{QLoRA} \\ \cmidrule(lr){2-4} \cmidrule(lr){5-7} \cmidrule(lr){8-10} \cmidrule(lr){11-13} \cmidrule(lr){14-14} \cmidrule(lr){15-15}
Metric                     & ROUGE & BGE & LAJ & ROUGE & BGE & LAJ & ROUGE & BGE & LAJ & ROUGE & BGE & LAJ & Accuracy & Accuracy\\ \midrule
Initial                & 28.53  & 74.05 & 2.94  & 34.42  & 78.54  & 3.00  & 24.85 & 72.45  & 2.26  & 34.29  & 80.82  & 3.03 & 70.7 & 70.8 \\
Random                 & 40.07  & 84.04 & 3.26  & 41.68  & 84.26  & 3.22  & 36.95 & 80.47  & 3.12  & 38.64  & 80.46  & 3.07 & 71.9 & 71.9 \\
\cmidrule{1-1}
SelectIT               & 46.51  & 86.18 & 3.25  & 50.31  & 87.38  & 3.25  & 41.42 & 83.25  & 3.27  & 44.51  & 84.18  & 3.34 & 72.7 & 73.0 \\
+ DistilGPT2           & 41.26  & 80.33 & 3.20  & 44.86  & 84.72  & 3.23  & 39.18 & 80.99  & 2.99  & 41.72  & 81.50  & 3.14 & 72.0 & 72.7 \\
+ \sysn{}              & 46.48  & 85.86 & 2.28  & 50.87  & 87.43  & 3.26  & 42.07 & 83.67  & 3.27  & 44.99  & 85.13  & 3.37 & 74.7 & 72.9 \\
\cmidrule{1-1}
LESS                   & 48.21  & 86.19 & 3.34  & 51.24  & 86.07  & 3.37  & 43.34 & 84.19  & 3.38  & 44.73  & 84.04  & 3.32 & 75.6 & 76.7 \\
+ DistilGPT2           & 42.18  & 78.34 & 3.23  & 48.64  & 79.09  & 3.27  & 42.02 & 80.89  & 3.29  & 42.51  & 82.35  & 3.29 & 74.6 & 74.1 \\
+ \sysn{}              & 48.20  & 86.31 & 3.36  & 51.56  & 86.39  & 3.41  & 44.42 & 84.69  & 3.32  & 46.40  & 85.44  & 3.36 & 75.0 & 76.5 \\
\cmidrule{1-1}
DELIFT (SE)            & 48.36  & 85.91 & 3.38  & 51.43  & 86.20  & 3.34  & 44.30 & 85.52  & 3.41  & 45.35  & 86.34  & 3.48 & 78.9 & 79.8 \\
+ DistilGPT2           & 47.21  & 84.24 & 3.28  & 49.37  & 84.24  & 3.29  & 43.51 & 85.45  & 3.41  & 44.89  & 79.81  & 3.36 & 75.4 & 76.1 \\
+ \sysn{}              & 48.59  & 85.01 & 3.39  & 50.53  & 86.10  & 3.33  & 45.49 & 86.27  & 3.44  & 45.75  & 86.45  & 3.47 & 78.5 & 79.8 \\ 
\cmidrule{1-1}
DELIFT                 & 51.66  & 88.02 & 3.43  & 55.58  & 91.81  & 3.50  & 46.49 & 87.60  & 3.50  & 49.16  & 87.74  & 3.54 & 81.5 & 83.1 \\
+ DistilGPT2           & 47.09  & 84.74 & 3.26  & 48.21  & 84.24  & 3.28  & 45.08 & 81.45  & 3.41  & 41.07  & 83.22  & 3.44 & 77.1 & 78.5 \\
+ \sysn{}              & 52.03  & 88.38 & 3.41  & 55.85  & 91.96  & 3.51  & 46.26 & 87.41  & 3.55  & 49.15  & 87.74  & 3.50 & 82.0 & 83.6 \\
\midrule
Full Data              & 54.43  & 92.55 & 3.40  & 59.47  & 94.12  & 3.58  & 48.53 & 91.21  & 3.63  & 48.29  & 90.82  & 3.66 & 80.5 & 81.6 \\
\bottomrule
\end{tabular}
}
\caption{Results on the Llama-8B model with $v=0.3, u=0.05$. ``+ \sysn{}'' indicates using \sysn{} to estimate influence values computed from the corresponding method's influence function. ``+ DistilGPT2'' indicates using the DistilGPT2 model as the language model in the corresponding method's influence function. The average performance difference between \sysn{} and the original influence function is merely 0.13\%.}
\label{table: llama v=0.3}
\end{table*}

\begin{table*}[th]
\centering\scriptsize
\makebox[\linewidth][c]{%
\begin{tabular}{lcccccccccccccc}
\toprule
Dataset                 & \multicolumn{6}{c}{MixInstruct}                                                                                                         & \multicolumn{6}{c}{Alpaca}   & \multicolumn{2}{c}{MMLU}                                                                                                           \\ \cmidrule(lr){2-7} \cmidrule(lr){8-13} \cmidrule{14-15}
Method                & \multicolumn{3}{c}{ICL} & \multicolumn{3}{c}{QLoRA} & \multicolumn{3}{c}{ICL} & \multicolumn{3}{c}{QLoRA} & \multicolumn{1}{c}{ICL} & \multicolumn{1}{c}{QLoRA} \\ \cmidrule(lr){2-4} \cmidrule(lr){5-7} \cmidrule(lr){8-10} \cmidrule(lr){11-13} \cmidrule(lr){14-14} \cmidrule(lr){15-15}
Metric                     & ROUGE & BGE & LAJ & ROUGE & BGE & LAJ & ROUGE & BGE & LAJ & ROUGE & BGE & LAJ & Accuracy & Accuracy\\ \midrule
Initial      & 16.19  & 61.32  & 2.06  & 19.31   & 64.27   & 2.09  & 24.53  & 71.42  & 2.48  & 24.80   & 71.79   & 2.61  & 69.4 & 70.9\\
Random       & 28.33  & 72.41  & 2.37  & 29.93   & 75.78   & 2.50  & 26.54  & 72.71  & 2.66  & 28.10   & 73.00   & 2.78  & 68.9 & 71.9 \\ \cmidrule{1-1}
SelectIT     & 40.64  & 72.63  & 2.21  & 44.85   & 75.72   & 2.83  & 31.86  & 76.29  & 2.73  & 32.56   & 78.17   & 2.77  & 71.7 & 71.6 \\
+ DistilGPT2 & 40.33  & 71.49  & 2.11  & 43.26   & 74.14   & 2.38  & 29.65  & 75.32  & 2.62  & 31.66   & 74.20   & 2.67  & 69.9 & 70.8 \\
+ \sysn{}    & 41.59  & 71.36  & 2.26  & 45.87   & 74.22   & 2.58  & 32.05  & 76.82  & 2.78  & 32.90   & 77.12   & 2.80  & 71.1 & 72.0 \\ \cmidrule{1-1}
LESS         & 45.33  & 78.68  & 3.03  & 46.03   & 81.04   & 3.05  & 38.43  & 78.83  & 2.98  & 41.68   & 81.83   & 3.09  & 74.0 & 75.6 \\
+ DistilGPT2 & 43.15  & 74.69  & 2.42  & 42.87   & 75.80   & 2.46  & 34.76  & 75.26  & 2.86  & 37.66   & 78.57   & 3.06  & 66.7 & 69.3 \\
+ \sysn{}    & 46.32  & 78.84  & 3.05  & 47.84   & 80.48   & 3.03  & 39.59  & 78.75  & 2.89  & 42.06   & 81.06   & 3.08  & 74.1 & 74.9 \\ \cmidrule{1-1}
DELIFT (SE)  & 46.68  & 81.01  & 3.12  & 48.42   & 83.67   & 3.15  & 42.52  & 81.21  & 3.12  & 43.83   & 84.35   & 3.26  & 74.0 & 75.1 \\
+ DistilGPT2 & 45.89  & 79.77  & 3.07  & 46.07   & 80.40   & 3.10  & 41.15  & 79.42  & 2.96  & 42.32   & 82.63   & 3.02  & 75.0 & 75.4 \\
+ \sysn{}    & 46.81  & 81.23  & 3.14  & 48.64   & 82.76   & 3.16  & 42.72  & 80.86  & 3.16  & 42.75   & 84.52   & 3.27  & 75.2 & 75.5 \\ \cmidrule{1-1}
DELIFT       & 48.85  & 83.89  & 3.25  & 50.90   & 85.64   & 3.27  & 44.74  & 83.60  & 3.28  & 46.33   & 87.87   & 3.46  & 77.4 & 79.4 \\
+ DistilGPT2 & 43.69  & 77.83  & 3.04  & 45.61   & 79.76   & 3.06  & 40.25  & 77.10  & 2.96  & 43.37   & 79.88   & 2.96  & 74.6 & 75.4 \\
+ \sysn{}    & 48.97  & 83.57  & 3.25  & 49.71   & 86.45   & 3.29  & 45.78  & 85.49  & 3.29  & 48.69   & 87.14   & 3.48  & 77.5 & 79.3 \\ \midrule
Full Data    & 49.31  & 86.25  & 3.48  & 52.55   & 89.58   & 3.51  & 49.31  & 89.39  & 3.45  & 50.95   & 90.23   & 3.66  & 73.7 & 74.6 \\  
\bottomrule
\end{tabular}
}
\caption{Results on the Qwen2.5 with $v=0.3, u=0.05$. ``+ \sysn{}'' indicates using \sysn{} to estimate influence values computed from the corresponding method's influence function. ``+ DistilGPT2'' indicates using the DistilGPT2 model as the language model in the corresponding method's influence function. The average performance difference between \sysn{} and the original influence function is merely 0.24\%.}
\label{table: qwen v=0.3}
\end{table*}

\begin{table*}[th]
\centering\scriptsize
\makebox[\linewidth][c]{%
\begin{tabular}{lcccccccccccccc}
\toprule
Dataset                 & \multicolumn{6}{c}{MixInstruct}                                                                                                         & \multicolumn{6}{c}{Alpaca}   & \multicolumn{2}{c}{MMLU}                                                                                                           \\ \cmidrule(lr){2-7} \cmidrule(lr){8-13} \cmidrule{14-15}
Method                & \multicolumn{3}{c}{ICL} & \multicolumn{3}{c}{QLoRA} & \multicolumn{3}{c}{ICL} & \multicolumn{3}{c}{QLoRA} & \multicolumn{1}{c}{ICL} & \multicolumn{1}{c}{QLoRA} \\ \cmidrule(lr){2-4} \cmidrule(lr){5-7} \cmidrule(lr){8-10} \cmidrule(lr){11-13} \cmidrule(lr){14-14} \cmidrule(lr){15-15}
Metric                     & ROUGE & BGE & LAJ & ROUGE & BGE & LAJ & ROUGE & BGE & LAJ & ROUGE & BGE & LAJ & Accuracy & Accuracy\\ \midrule
Initial      & 32.31  & 74.27  & 2.18  & 35.16   & 78.13   & 2.21  & 37.44  & 78.65  & 2.41  & 37.81   & 78.69   & 2.43  & 71.9        & 72.3       \\
Random       & 35.83  & 80.62  & 2.98  & 37.13   & 81.57   & 3.01  & 40.44  & 82.75  & 2.48  & 40.19   & 81.15   & 2.46  & 71.3        & 72.3       \\ \cmidrule{1-1}
SelectIT     & 40.64  & 84.99  & 3.17  & 45.97   & 86.31   & 3.20  & 42.72  & 82.00  & 2.54  & 43.06   & 83.59   & 2.67  & 73.6        & 75.8       \\
+ DistilGPT2 & 39.84  & 80.48  & 3.02  & 41.81   & 81.21   & 3.03  & 40.74  & 81.44  & 2.37  & 40.36   & 81.85   & 2.53  & 73.5        & 73.8       \\
+ \sysn{}    & 40.33  & 84.95  & 3.14  & 46.85   & 86.06   & 3.19  & 41.52  & 82.18  & 2.57  & 43.45   & 82.86   & 2.63  & 73.6        & 75.4       \\ \cmidrule{1-1}
LESS         & 48.33  & 85.78  & 3.30  & 49.93   & 87.42   & 3.37  & 44.14  & 85.80  & 3.04  & 47.13   & 87.94   & 3.05  & 75.4        & 76.6       \\
+ DistilGPT2 & 45.22  & 82.80  & 3.23  & 44.53   & 84.56   & 3.31  & 40.92  & 80.23  & 2.68  & 42.65   & 84.29   & 2.96  & 74.5        & 74.4       \\
+ \sysn{}    & 48.75  & 86.15  & 3.33  & 51.93   & 86.57   & 3.41  & 44.43  & 85.79  & 3.07  & 47.90   & 88.01   & 3.04  & 75.2        & 77.3       \\ \cmidrule{1-1}
DELIFT (SE)  & 48.84  & 88.10  & 3.42  & 48.85   & 88.16   & 3.44  & 45.13  & 87.34  & 3.35  & 48.47   & 89.19   & 3.42  & 76.7        & 77.1       \\
+ DistilGPT2 & 46.25  & 86.22  & 3.39  & 46.29   & 87.67   & 3.41  & 44.84  & 86.00  & 3.25  & 46.86   & 86.93   & 3.38  & 75.7        & 77.7       \\
+ \sysn{}    & 48.81  & 88.77  & 3.41  & 49.14   & 88.33   & 3.44  & 46.01  & 86.43  & 3.36  & 47.99   & 88.03   & 3.41  & 76.8        & 77.4       \\ \cmidrule{1-1}
DELIFT       & 53.70  & 91.52  & 3.54  & 54.69   & 92.42   & 3.56  & 50.36  & 89.53  & 3.40  & 53.65   & 91.79   & 3.53  & 78.2        & 80.4       \\
+ DistilGPT2 & 46.55  & 88.81  & 3.41  & 47.78   & 89.34   & 3.46  & 45.31  & 83.69  & 3.23  & 44.78   & 85.08   & 3.31  & 75.3        & 76.9       \\
+ \sysn{}    & 53.53  & 90.82  & 3.52  & 56.65   & 91.42   & 3.55  & 50.67  & 88.97  & 3.43  & 52.48   & 92.03   & 3.53  & 78.4        & 80.9       \\ \midrule
Full Data    & 52.30  & 91.32  & 3.60  & 54.83   & 91.57   & 3.69  & 50.61  & 90.77  & 3.41  & 54.51   & 90.70   & 3.58  & 79.1        & 81.2       \\ 
\bottomrule
\end{tabular}
}
\caption{Results on the Mistral model with $v=0.3, u=0.05$. ``+ \sysn{}'' indicates using \sysn{} to estimate influence values computed from the corresponding method's influence function. ``+ DistilGPT2'' indicates using the DistilGPT2 model as the language model in the corresponding method's influence function. The average performance difference between \sysn{} and the original influence function is merely 0.12\%.}
\label{table: mistral v=0.3}
\end{table*}

\subsection{Interpretive Analysis}
\label{subsec: interpretive_analysis}
\begin{figure*}[h]
    \centering
    \includegraphics[width=0.3\linewidth]{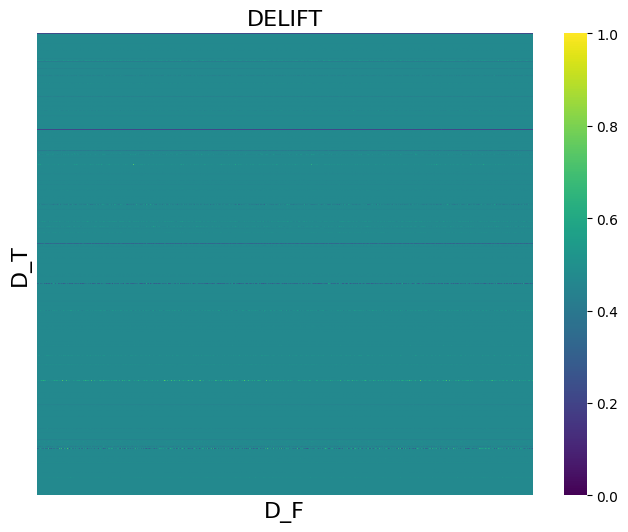}
    \includegraphics[width=0.3\linewidth]{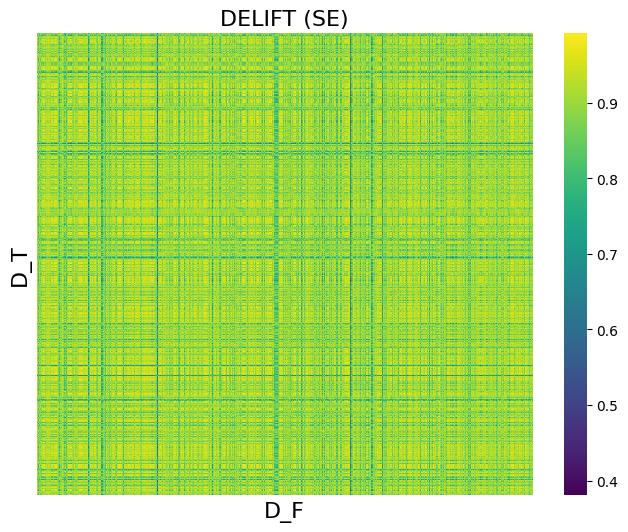}
    \includegraphics[width=0.3\linewidth]{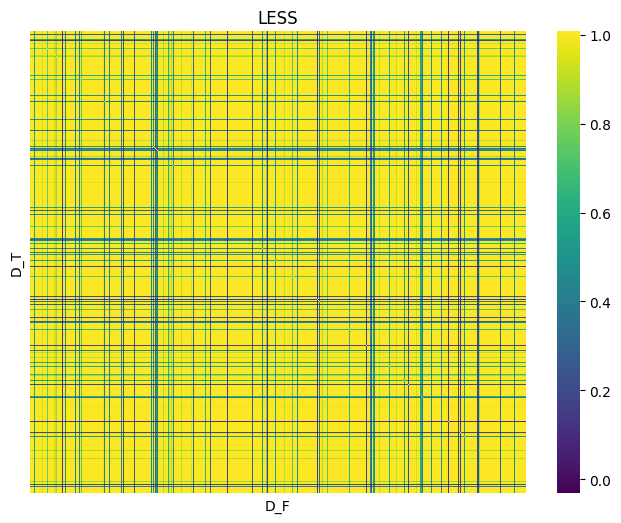}
    \caption{Distribution of influence values across each of the methods on the Alpaca dataset. To clarify, "D\_F" is $\mathcal{D_F}$ and "D\_T" is $\mathcal{D_T}$. The x-axis spans the 15,000 examples from $\mathcal{D_F}$ and the y-axis spans the 5,000 samples from $\mathcal{D_T}$. SelectIT only has an x-axis.}
    \label{fig: influence_distribution}
\end{figure*}

We posit two reasons that contribute to the InfluenceNetwork's success: (1) most data points have the same influence, and (2) influence estimation is a lossy task. 

Firstly, we visualize the distribution of the influence values in Figure \ref{fig: influence_distribution}. As shown, the most "different" influence values (the stark dark/light colors indicating low/high influence) are sparse. This indicates that \textit{the neural network simply has to learn to estimate the extremes}, and can achieve good performance for the rest of the values. Second, we notice that \textit{influence estimation is an incredibly lossy task}. It involves compressing high-dimensional text representations to a singular, scalar value. Furthermore, the scalar values are constrained to a small range (see Equation \ref{equation: influence_net_formulation}), reducing the margin of error. Putting these two reasons together, we can see why a neural network can replace a language model during influence estimation without any significant effect on performance.

\section{Subset Selection Evaluation}
\label{sec: subset_selection_evaluation}
Motivated by the results in Figure \ref{fig: influence_net}, we apply the InfluenceNetwork to the downstream task of subset selection: \textit{can we achieve the same performance when using the InfluenceNetwork instead of the original influence function}? Thus, this section corresponds to Step 3 in Figure \ref{fig: nncift}.

\paragraph{Datasets and models.} We use MixInstruct, \citep{alpaca}, Alpaca \citep{alpaca}, and MMLU \citep{mmlu} to evaluate \sysn{}. These are instruction-tuning, preference alignment, and knowledge-based benchmarks where we use 15k for training, 5k for validation, and 5k for testing. We evaluate using three models: \texttt{meta-llama/Llama-3.1-8B} \citep{grattafiori2024llama3herdmodels}, \texttt{Qwen/Qwen2.5-1.5B} \citep{qwen2.5}, and \texttt{mistralai/Mistral-Small-Instruct-2409} (22.2B) \citep{mistral_small}. Note, we use Llama-8b, Qwen2.5 and Mistral as shorthand for the rest of the experimental section.

\paragraph{Metrics.} To evaluate the instruction following capabilities of our fine-tuned model $\mathcal{M}'$, we employ a variety of metrics to capture the similarity between ground truth answers and predicted answers from $\mathcal{M}'$: (1) ROUGE \citep{rouge}: $n$-gram word overlap (specifically, rouge-1), (2) BGE: semantic similarity of embeddings using \texttt{bge-large-en-v1.5}, and (3) LAJ: an LLM-as-a-Judge, namely the \texttt{prometheus-7b-v2.0} model \citep{prometheus}. Prometheus' grading rubric is borrowed from \citet{delift} in Appendix B. Next, to evaluate the costs of each method, we use time (in seconds) took on 2 Nvidia A40 GPUs. However, for MMLU, we use classification accuracy.

\paragraph{Baselines.} Besides the influence functions DELIFT, DELIFT (SE), LESS, and SelectIT, we include three other baselines: Initial, DistilGPT2, and Full Data. \textit{Initial} is the setting where $v=0.0$. This is the base model's performance on the dataset. Next, we use a small language model \textit{DistilGPT2} (\texttt{distilbert/distilgpt2}) \citep{distilgpt2} which has 88.2M parameters as the underlying language/embedding model in the influence functions. Finally, \textit{Full Data} is the setting where $v=1.0$, i.e., the model's performance when the full dataset is used.

\paragraph{Setup.}
We use $u=0.05$ for training the InfluenceNetwork. We also use a small fraction of $\mathcal{D_F}$ to fine-tune the language model -- we call this fraction $v$. We evaluate with $v=0.3$. Our evaluation framework includes two different settings to fine-tune the language model: using the selected subset of data points as (1) PEFT data for QLoRA \citep{qlora} on $\mathcal{M}$, or (2) in-context learning (ICL) examples. To elaborate on the ICL set up, we choose the top-5 most semantically similar samples from the chosen subset to add in-context. To measure semantic similarity, we again use \texttt{bge-large-en-v1.5}. Table \ref{table: llama v=0.3}-\ref{table: mistral v=0.3} reports results for each model on all three datasets with $v=0.3$; Table \ref{table: actual_costs} reports the cost in time for each method. All tables report the results for one run.

\subsection{Analysis}

Table \ref{table: actual_costs} reports the costs for each method, in seconds. It shows that \textbf{data valuation can be performed at 77-99\% faster} than the original influence functions. This is because the number of parameters in \sysn{} is 0.00096-0.013\% the size of the language model in the original influence function. Also, when using the DistilGPT2 model, which is near 1\% the size of the language model, the costs are reduced by 54-91\%. While these results are promising, the results on the downstream task of subset selection clearly differentiate \sysn{} and the DistilGPT2 baseline. \textbf{Despite the significant speedups, \sysn{} shows no compromise to performance}, as shown in Tables \ref{table: llama v=0.3}-\ref{table: mistral v=0.3}.

To begin, the pairwise functions outperform the pointwise function (SelectIT) because they are able to capture more fine-grained effects of the data point on a model's learning. Next, DELIFT and DELIFT (SE) are able to outperform LESS because the theoretical guarantees of using submodular functions yields improved empirical performance. Finally, DELIFT uses model dependent information, tailoring the subset to the model's weaknesses, allowing it to outperform DELIFT (SE).

Keeping these in mind, \textbf{\sysn{} is able to achieve performance comparable to the original data valuation methods, even across models and datasets}. DistilGPT2 shows performance degradations, especially in the model-dependent methods (DELIFT, LESS, and SelectIT). This is because the model-dependent methods experience significant performance gains when the data valuation model is the same as the fine-tuning model. We note that our evaluation's focus is that \sysn{} works as well as the original influence functions, and not the comparison of performance between them.

The absolute average performance difference across metrics between the original influence functions and \sysn{} is only 0.16\%\footnote{The average performance difference is calculated by taking the absolute difference in performance, dividing it by the original performance, and then averaging this ratio across all settings (datasets, methods, metrics, baselines).}. Because the neural network is able to estimate the influence values with great accuracy, the selected subsets of data would be mostly the same between the original influence function and \sysn{}. Hence, the performance difference of 0.16\% can be attributed as the variability in the language model's performance between two runs. Additionally, this trend is consistent across datasets and models, which shows the wide applicability of our method.

\begin{table}
\centering
\resizebox{0.8\columnwidth}{!}{
\begin{tabular}{lcccc}
\toprule
Model                  & \multicolumn{2}{c}{Phi-3} & \multicolumn{2}{c}{Llama-8B}  \\ \cmidrule(lr){2-3} \cmidrule(lr){4-5}
Dataset                & MixInstruct & Alpaca & MixInstruct & Alpaca \\ \midrule
Initial                & - & -  & - & - \\
Random                 & 12.4 & 12.3 & 12.9 & 12.3 \\ 
\cmidrule{1-1}
SelectIT               & 7,047 & 6,594 & 6,671 & 6,470 \\
DistilGPT2 + SelectIT  & 144 & 139  & 144  & 139        \\
\sysn{} + SelectIT     & 65 & 63 & 64 & 63 \\ 
\cmidrule{1-1}
LESS                   & 12,338 & 11,217 & 10,843 & 14,819 \\
DistilGPT2 + LESS      & 1,291 & 1,278  & 1,291  & 1,278        \\
\sysn{} + LESS         & 78 & 75 & 74 & 84 \\ 
\cmidrule{1-1}
DELIFT (SE)            & 216 & 218 & 218 & 219 \\
DistilGPT2 + DELIFT(SE)& 98 & 99  & 98  & 99        \\
\sysn{} + DELIFT (SE)  & 48 & 48 & 48 & 48 \\ 
\cmidrule{1-1}
DELIFT                 & 67,379 & 68,117 & 68,076 & 65,711 \\
DistilGPT2 + DELIFT    & 8,058 & 7,790  & 8,058  & 7,790        \\
\sysn{} + DELIFT & 215 & 217 & 217 & 211 \\
\cmidrule{1-1}
Full Data              & - & - & - & -  \\
\bottomrule \\
\end{tabular}
}
\caption{Costs (in seconds) of data valuation. Specifically: \textbf{Random}: chooses a random subset of points. \textbf{SelectIT}: calculates the ranking scores for each data point according to Appendix \ref{app: pointwise_influence_functions}. \textbf{LESS}: computes the cosine similarity between pairs of projected gradients for $\mathcal{D_F}$ and $\mathcal{D_T}$, according to Equation \ref{less_influence_function}. \textbf{DELIFT (SE)}: computes the distance between each pair of embeddings $(i, j): i \in \mathcal{D_F}, j \in \mathcal{D_T}$, according to Equation \ref{delift_se_influence_function}. \textbf{DELIFT}: computes the inference-based utility metric for each pair of embeddings $(i, j)$, according to Equation \ref{delift_influence_function}. \textbf{\sysn{}}: Steps 1 and 2 in Figure \ref{fig: nncift}. Note, the costs of DistilGPT2 are the same across both models because they use the same data valuation.
}
\label{table: actual_costs}
\end{table}

\section{Conclusion}
In this paper, we introduce \sysn{}: \textbf{N}eural \textbf{N}etworks for effi\textbf{C}ient \textbf{I}nstruction \textbf{F}ine-\textbf{T}uning to distill highly parameterized models used in modern influence functions into small neural networks. We empirically show the effectiveness of our InfluenceNetwork design through low prediction error rates, and competitive performance on the downstream task of subset selection for IFT. We use four different influence functions to test with \sysn{}; our experimentation shows that \sysn{} can lower costs for expensive data valuation, is adaptive to all kinds of influence functions (model-dependent or -independent; pairwise or pointwise), and does not require retraining for new data. Future work will focus on two things: applicability and more targeted modeling. To improve applicability, incorporate more fine-tuning stage objectives such as task-specific dataset selection or continual learning. To target the modeling, we can shift from learning influence between data points to estimating influence between data and a model parameters. Finally, we leave to future work to generalize \sysn{} to distributional shifts within the data (although, the training is lightweight enough that retraining would suffice for now).

\bibliography{references.bib}
\bibliographystyle{abbrvnat}

\appendix

\section{Hyperparameter Studies}
\begin{figure*}[h]
    \centering
    \includegraphics[width=\linewidth]{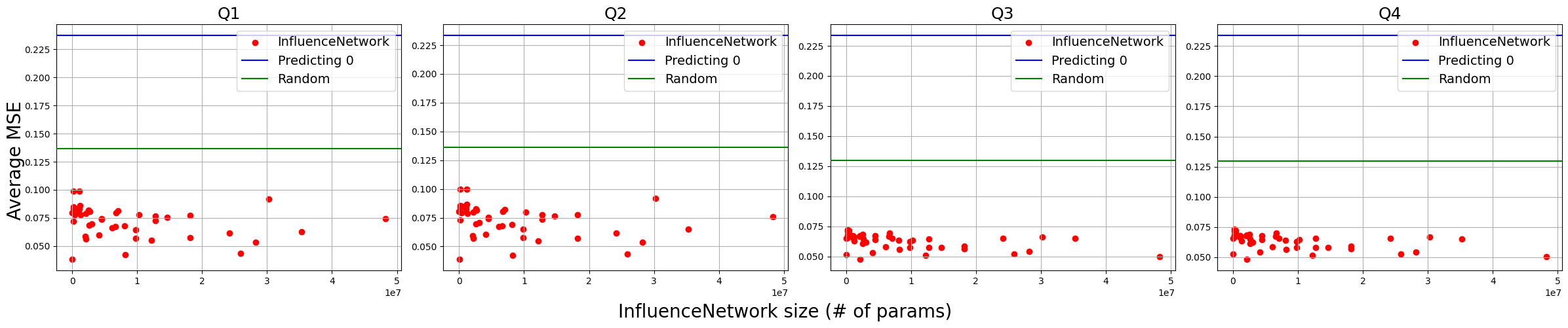}
    \caption{MSE versus InfluenceNetwork sizes (measured by the number of parameters). We try 1-5 layers with 46 different combinations of hidden layer sizes from \{5, 10, 20, 50, 100, 200, 500, 1000, 2000, 3000, 4000, 5000\}.}
    \label{fig: influence_network_size}
\end{figure*}

\subsection{Hyperparameter Study \#1: InfluenceNetwork sizes}
\label{subsec: in_size}
We vary the number of layers and dimensions of each layer. For simplicity, we plot the number of parameters in the InfluenceNetwork versus the MSE. The results can be found in Figure \ref{fig: influence_network_size}. This figure shows that small InfluenceNetwork's perform comparatively well as larger InfluenceNetwork's.

\subsection{Hyperparameter study \#2: Trade-off between $u$ and $v$}
\label{subsec: hp_u_v}

We perform a hyperparameter study between $u$ and $v$ on MixInstruct using DELIFT's influence function (Equation \ref{delift_influence_function}). We perform a grid search where $u = v = \{0, 0.01, 0.05, \allowbreak 0.1, 0.15, 0.20, 0.25, 0.3, 0.4, 0.5, 0.6, 0.7, 0.8\}$, amounting to 169 experiments. Figure \ref{fig: hyperparam_study} shows the results using the BGE metric from each of these experiments. As shown, the two figures in each row follow the same general trend, showcasing that \textbf{\sysn{} can effectively replace the expensive influence function estimation}.

\begin{wraptable}{r}{8cm}
    \centering
    \includegraphics[width=0.45\linewidth]{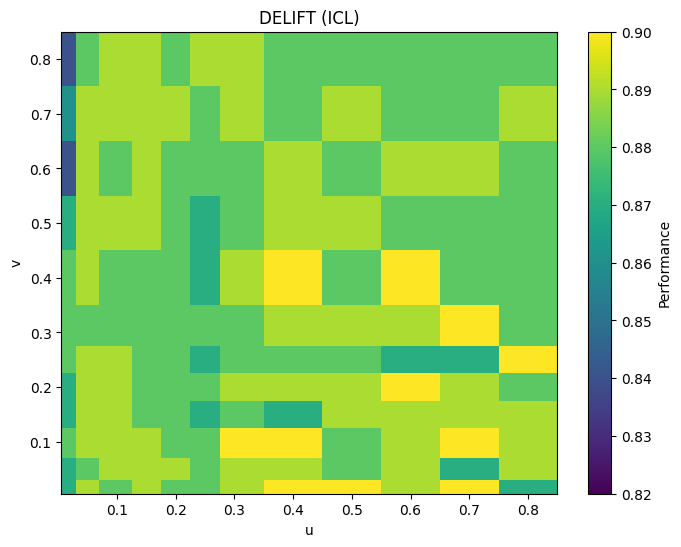}
    \includegraphics[width=0.45\linewidth]{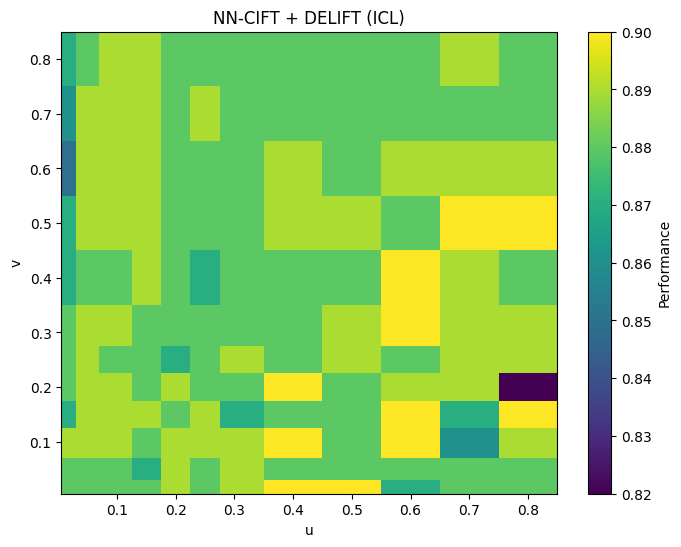}
    \includegraphics[width=0.45\linewidth]{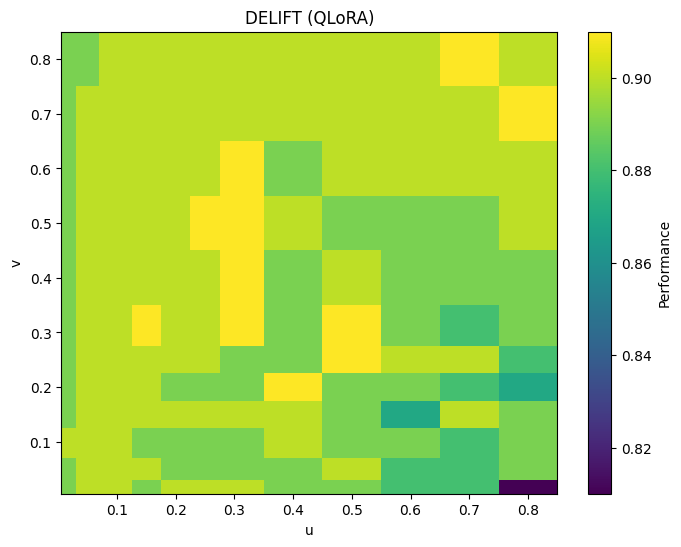}
    \includegraphics[width=0.45\linewidth]{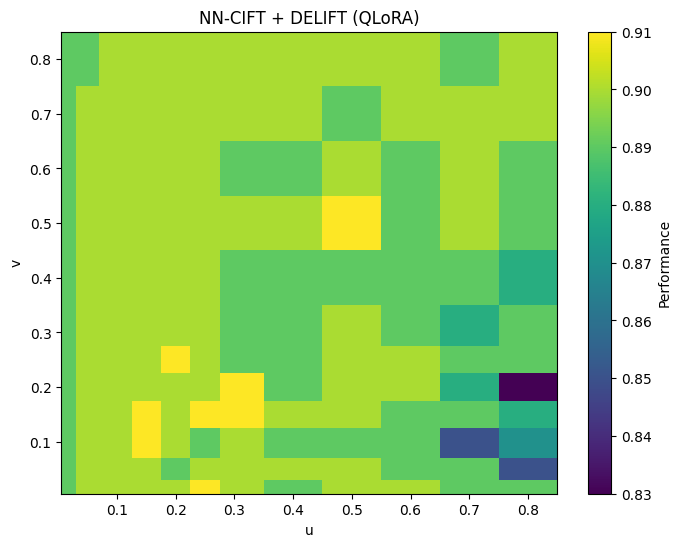}
    \caption{Hyperparameter study for $u$ and $v$ on MixInstruct with DELIFT's influence function. Lighter colors indicate better BGE performance.}
    \label{fig: hyperparam_study}
\end{wraptable}


As expected, we notice a few trends. (1) QLoRA generally has better performance than ICL. This is because fine-tuning has more impact on the model than simply adding examples to the prompt (i.e., prompt engineering). (2) The bottom right tends to be darker as fewer IFT data lead to insufficient training. (3) Larger IFT subsets, especially in the ICL setting, lead to poorer performance. During ICL, the top-5 semantically similar samples are chosen from the subset to add as in-context examples. However, semantic similarity does not always translate to performance enhancement as these samples can be harmful to the model's performance. Finally, a follow-up to (3), the highest performance regions tend to be around $v=0.2$ - $0.4$. Appendix \ref{app: smaller_subset_eval} contains results on smaller subsets of IFT data ($v=0.1$ and $0.2$).

\section{Evaluation on Smaller Subsets}
\label{app: smaller_subset_eval}
Tables \ref{table: phi v=0.1} and \ref{table: phi v=0.2} report extra results for the Phi-3 model on $v=0.1$ and $v=0.2$, respectively. Similarly, Tables \ref{table: llama v=0.1} and \ref{table: llama v=0.2} report results for Llama-8B on $v=0.1$ and $v=0.2$, respectively. With Table \ref{table: llama v=0.3} in the main text, these results show an increasing trend in performance with a higher subset of IFT data (i.e., higher $v$). They also show similar trends where \sysn{} performs similarly to the original influence function.

\begin{table*}[t]
\centering\scriptsize
\makebox[\linewidth][c]{%
\begin{tabular}{lcccccccccccc}
\toprule
Dataset                 & \multicolumn{6}{c}{MixInstruct}                                                                                                         & \multicolumn{6}{c}{Alpaca}                                                                                                              \\ \cmidrule(lr){2-7} \cmidrule(lr){8-13} 
Method                & \multicolumn{3}{c}{ICL}                                            & \multicolumn{3}{c}{QLoRA}                                          & \multicolumn{3}{c}{ICL}                                            & \multicolumn{3}{c}{QLoRA}                                          \\ \cmidrule(lr){2-4} \cmidrule(lr){5-7} \cmidrule(lr){8-10} \cmidrule(lr){11-13}
Metric                     & ROUGE                & BGE                  & LAJ                  & ROUGE                & BGE                  & LAJ                  & ROUGE                & BGE                  & LAJ                  & ROUGE                & BGE                  & LAJ                  \\ \midrule
Initial                & 37.87  & 78.92  & 2.98   & 36.36  & 82.55  & 3.02  & 25.79 & 67.82  & 2.56  & 27.29  & 71.57  & 2.62       \\
Random                 & 37.51  & 78.01  & 3.05   & 35.55  & 82.13  & 3.04  & 24.33 & 67.37  & 2.84  & 29.34  & 70.86  & 3.06       \\ 
\cmidrule{1-1}
SelectIT               & 33.20  & 72.12  & 3.12    & 37.00  & 73.45  & 3.13 & 24.48 & 67.48  & 2.86  & 30.06  & 68.06  & 3.04       \\
\sysn{} + SelectIT     & 33.55  & 72.15  & 3.07    & 35.38  & 72.45  & 3.18 & 26.41 & 65.57  & 2.81  & 28.78  & 67.83  & 2.99       \\ 
\cmidrule{1-1}
LESS                   & 32.57  & 72.07  & 3.05    & 34.61  & 72.82  & 3.18 & 26.15 & 69.83  & 2.81  & 28.53  & 67.17  & 2.99       \\
\sysn{} + LESS         & 33.19  & 72.94  & 3.02    & 35.42  & 72.03  & 3.18 & 24.63 & 70.11  & 2.84  & 27.63  & 67.41  & 2.51       \\ 
\cmidrule{1-1}
DELIFT (SE)            & 35.71  & 78.09  & 3.22    & 39.63  & 78.36  & 3.28 & 29.17 & 70.69  & 3.01  & 30.60  & 71.50  & 3.14       \\
\sysn{} + DELIFT (SE)  & 36.34  & 78.02  & 3.22    & 39.75  & 78.76  & 3.33 & 29.22 & 72.28  & 3.03  & 30.23  & 71.01  & 3.16       \\ 
\cmidrule{1-1}
DELIFT                 & 36.45  & 78.11  & 3.23    & 39.83  & 78.83  & 3.29 & 30.15 & 74.01  & 3.18  & 37.81  & 78.49  & 3.31       \\
\sysn{} + DELIFT       & 36.17  & 78.16  & 3.22    & 38.08  & 78.25  & 3.28 & 31.95 & 74.84  & 3.26  & 37.26  & 78.36  & 3.28       \\ 
\midrule
Full Data              & 58.65  & 88.72  & 3.45    & 65.51  & 92.24  & 3.51 & 35.27 & 77.85  & 3.31  & 39.29  & 78.85  & 3.29       \\
\bottomrule
\end{tabular}
}
\caption{Results on the Phi-3 model with $v=0.1, u=0.05$. \sysn{} + Method and DistilGPT2 + Method follow the same definitions as in Table \ref{table: mistral v=0.3}. The average performance difference between \sysn{} and the original influence function is merely 1.91\%.}
\label{table: phi v=0.1}
\end{table*}

\begin{table*}[t]
\centering\scriptsize
\makebox[\linewidth][c]{%
\begin{tabular}{lcccccccccccc}
\toprule
Dataset                 & \multicolumn{6}{c}{MixInstruct}                                                                                                         & \multicolumn{6}{c}{Alpaca}                                                                                                              \\ \cmidrule(lr){2-7} \cmidrule(lr){8-13} 
Method                & \multicolumn{3}{c}{ICL}                                            & \multicolumn{3}{c}{QLoRA}                                          & \multicolumn{3}{c}{ICL}                                            & \multicolumn{3}{c}{QLoRA}                                          \\ \cmidrule(lr){2-4} \cmidrule(lr){5-7} \cmidrule(lr){8-10} \cmidrule(lr){11-13}
Metric                     & ROUGE                & BGE                  & LAJ                  & ROUGE                & BGE                  & LAJ                  & ROUGE                & BGE                  & LAJ                  & ROUGE                & BGE                  & LAJ                  \\ \midrule
Initial                & 28.53  & 74.05 & 2.94  & 34.42  & 78.54  & 3.00  & 24.85 & 72.45  & 2.26  & 34.29  & 80.82  & 3.03       \\
Random                 & 35.67  & 76.30 & 3.18  & 37.20  & 80.63  & 3.19  & 30.82 & 75.38  & 2.82  & 36.95  & 80.48  & 3.05       \\
\cmidrule{1-1}
SelectIT               & 36.53  & 78.69 & 3.14  & 36.95  & 81.51  & 3.20  & 31.52 & 75.69  & 2.84  & 38.06  & 81.51  & 3.19       \\
\sysn{} + SelectIT     & 35.57  & 78.86 & 3.17  & 37.20  & 80.56  & 3.21  & 30.52 & 74.86  & 2.88  & 37.20  & 80.55  & 3.13       \\
\cmidrule{1-1}
LESS                   & 35.31  & 77.07 & 3.19  & 37.46  & 80.86  & 3.23  & 31.31 & 75.07  & 2.71  & 37.45  & 80.85  & 3.23       \\
\sysn{} + LESS         & 35.16  & 78.11 & 3.16  & 37.93  & 81.36  & 3.20  & 32.16 & 76.11  & 2.75  & 37.93  & 81.35  & 3.21       \\
\cmidrule{1-1}
DELIFT (SE)            & 35.13  & 77.71 & 3.12  & 36.78  & 79.69  & 3.15  & 30.14 & 73.71  & 2.61  & 36.80  & 79.69  & 3.15       \\
\sysn{} + DELIFT (SE)  & 35.12  & 78.69 & 3.13  & 37.33  & 80.34  & 3.08  & 31.12 & 74.69  & 2.62  & 37.33  & 80.34  & 3.08       \\ 
\cmidrule{1-1}
DELIFT                 & 37.82  & 80.55 & 3.18  & 37.61  & 82.63  & 3.20  & 31.82 & 75.62  & 2.83  & 37.61  & 80.55  & 3.29       \\
\sysn{} + DELIFT       & 37.52  & 81.02 & 3.15  & 37.88  & 82.01  & 3.19  & 31.55 & 75.04  & 2.79  & 37.88  & 81.16  & 3.29       \\
\midrule
Full Data              & 54.43  & 92.55 & 3.40  & 59.47  & 94.12  & 3.58  & 48.53 & 91.21  & 3.63  & 48.29  & 90.82  & 3.66       \\
\bottomrule
\end{tabular}
}
\caption{Results on the Llama-8b model with $v=0.1, u=0.05$. \sysn{} + Method and DistilGPT2 + Method follow the same definitions as in Table \ref{table: qwen v=0.3}. The average performance difference between \sysn{} and the original influence function is merely 1.14\%.}
\label{table: llama v=0.1}
\end{table*}
\begin{table*}[t]
\centering\scriptsize
\makebox[\linewidth][c]{%
\begin{tabular}{lcccccccccccccc}
\toprule
Dataset                 & \multicolumn{6}{c}{MixInstruct}                                                                                                         & \multicolumn{6}{c}{Alpaca}                                                                                                              \\ \cmidrule(lr){2-7} \cmidrule(lr){8-13} 
Method                & \multicolumn{3}{c}{ICL}                                            & \multicolumn{3}{c}{QLoRA}                                          & \multicolumn{3}{c}{ICL}                                            & \multicolumn{3}{c}{QLoRA}                                          \\ \cmidrule(lr){2-4} \cmidrule(lr){5-7} \cmidrule(lr){8-10} \cmidrule(lr){11-13}
Metric                     & ROUGE                & BGE                  & LAJ                  & ROUGE                & BGE                  & LAJ                  & ROUGE                & BGE                  & LAJ                  & ROUGE                & BGE                  & LAJ                  \\ \midrule
Initial                & 37.87  & 78.92  & 2.98   & 36.36  & 82.55  & 3.02  & 25.79 & 67.82  & 2.56  & 27.29  & 71.57  & 2.62       \\
Random                 & 37.91  & 78.96  & 3.06   & 38.89  & 81.88  & 3.05  & 29.95 & 76.35  & 3.12  & 30.27  & 76.21  & 3.15       \\ 
\cmidrule{1-1}
SelectIT               & 35.39  & 78.14  & 3.02    & 37.71  & 78.26  & 3.06 & 30.31 & 74.26  & 3.13  & 37.10  & 77.66  & 3.10       \\
\sysn{} + SelectIT & 35.71  & 78.23  & 3.04    & 37.36  & 78.24  & 3.05 & 31.03 & 75.79  & 3.09  & 36.67  & 77.98  & 3.04       \\ 
\cmidrule{1-1}
LESS                   & 37.61  & 79.55  & 3.07    & 37.43  & 78.93  & 3.09 & 32.57 & 74.07  & 3.02  & 34.61  & 76.68  & 3.08       \\
\sysn{} + LESS & 37.87  & 77.96  & 3.04    & 38.96  & 78.93  & 3.08 & 33.20 & 74.94  & 3.05  & 35.42  & 78.02  & 3.09       \\ 
\cmidrule{1-1}
DELIFT (SE)            & 39.56  & 81.25  & 3.17    & 39.77  & 82.74  & 3.15 & 34.06 & 77.31  & 3.23  & 39.48  & 80.95  & 3.25       \\
\sysn{} + DELIFT (SE) & 39.62  & 81.47  & 3.16    & 39.14  & 82.83  & 3.14 & 33.01 & 76.67  & 3.27  & 38.89  & 80.80  & 3.20       \\ 
\cmidrule{1-1}
DELIFT                 & 45.55  & 82.32  & 3.36    & 43.74  & 82.35  & 3.50 & 35.02 & 77.89  & 3.40  & 39.32  & 80.89  & 3.35       \\
\sysn{} + DELIFT & 46.44  & 82.47  & 3.38    & 43.76  & 82.72  & 3.52 & 34.44 & 77.39  & 3.36  & 38.30  & 80.32  & 3.31       \\ 
\midrule
Full Data              & 58.65  & 88.72  & 3.45    & 65.51  & 92.24  & 3.51 & 35.27 & 77.85  & 3.31  & 39.29  & 78.85  & 3.29       \\
\bottomrule
\end{tabular}
}
\caption{Results on the Llama-8b model with $v=0.2, u=0.05$. \sysn{} + Method and DistilGPT2 + Method follow the same definitions as in Table \ref{table: phi v=0.3}. The average performance difference between \sysn{} and the original influence function is merely 1.08\%.}
\label{table: phi v=0.2}
\end{table*}

\begin{table*}[t]
\centering\scriptsize
\makebox[\linewidth][c]{%
\begin{tabular}{lcccccccccccc}
\toprule
Dataset                 & \multicolumn{6}{c}{MixInstruct}                                                                                                         & \multicolumn{6}{c}{Alpaca}                                                                                                              \\ \cmidrule(lr){2-7} \cmidrule(lr){8-13} 
Method                & \multicolumn{3}{c}{ICL}                                            & \multicolumn{3}{c}{QLoRA}                                          & \multicolumn{3}{c}{ICL}                                            & \multicolumn{3}{c}{QLoRA}                                          \\ \cmidrule(lr){2-4} \cmidrule(lr){5-7} \cmidrule(lr){8-10} \cmidrule(lr){11-13}
Metric                     & ROUGE                & BGE                  & LAJ                  & ROUGE                & BGE                  & LAJ                  & ROUGE                & BGE                  & LAJ                  & ROUGE                & BGE                  & LAJ                  \\ \midrule
Initial                & 28.53  & 74.05 & 2.94  & 34.42  & 78.54  & 3.00  & 24.85 & 72.45  & 2.26  & 34.29  & 80.82  & 3.03       \\
Random                 & 39.55  & 82.79 & 3.25  & 39.05  & 82.64  & 3.26  & 31.49 & 76.96  & 3.06  & 41.67  & 79.77  & 3.14       \\
\cmidrule{1-1}
SelectIT               & 39.20  & 82.84 & 3.29  & 40.44 & 82.55   & 3.30  & 35.98 & 81.82  & 2.95  & 42.62  & 83.17  & 3.21       \\
\sysn{} + SelectIT     & 40.02  & 82.63 & 3.23  & 39.92 & 82.22   & 3.29  & 38.84 & 84.09  & 3.03  & 44.62  & 84.63  & 3.23       \\
\cmidrule{1-1}
LESS                   & 40.33  & 82.17 & 3.26  & 40.34  & 82.87  & 3.26  & 36.11 & 79.82  & 3.06  & 43.48  & 82.94  & 3.32       \\
\sysn{} + LESS         & 43.69  & 82.67 & 3.27  & 40.21  & 82.89  & 3.26  & 37.00 & 80.38  & 3.07  & 43.48  & 82.80  & 3.34       \\
\cmidrule{1-1}
DELIFT (SE)            & 44.57  & 82.63 & 3.31  & 45.97  & 83.87  & 3.33  & 38.52 & 82.37  & 3.18  & 45.73  & 83.33  & 3.35       \\
\sysn{} + DELIFT (SE)  & 45.03  & 83.69 & 3.30  & 45.97  & 83.95  & 3.40  & 38.57 & 82.18  & 3.17  & 45.20  & 82.79  & 3.39       \\ 
\cmidrule{1-1}
DELIFT                 & 45.55  & 83.69 & 3.37  & 48.21  & 86.81  & 3.36  & 39.16 & 82.30  & 3.26  & 45.24  & 83.38  & 3.39       \\
\sysn{} + DELIFT       & 46.40  & 84.73 & 3.34  & 47.81  & 86.83  & 3.31  & 40.16 & 82.37  & 3.28  & 45.67  & 83.49  & 3.41       \\
\midrule
Full Data              & 54.43  & 92.55 & 3.40  & 59.47  & 94.12  & 3.58  & 48.53 & 91.21  & 3.63  & 48.29  & 90.82  & 3.66       \\
\bottomrule
\end{tabular}
}
\caption{Results on the Llama-8b model with $v=0.2, u=0.05$. \sysn{} + Method and DistilGPT2 + Method follow the same definitions as in Table \ref{table: phi v=0.3}. The average performance difference between \sysn{} and the original influence function is merely 1.26\%.}
\label{table: llama v=0.2}
\end{table*}

\section{Influence Functions}
\label{app: influence_functions}
Following the problem formulation, we formally define the influence functions we used throughout our evaluation.

\subsection{Pairwise Influence Functions}
\label{app: pairwise influence functions}

\paragraph{DELIFT} \citep{delift} is a model-dependent, inference-based metric. Samples $(i_x,i_y) \in \mathcal{D_F}$ are used as in-context examples for evaluating $(j_x,j_y) \in \mathcal{D_T}$, and those with improved model performance are chosen to represent $\mathcal{D_T}$. This can be calculated by comparing the performance with and without $(i_x,i_y)$ as an in-context example (where $D(\cdot, \cdot) \in [0,1]$ is a function to measure distance between two probability distributions, and $f(q | \theta)$ is a language model with parameters $\theta$ and input query $q$):

\begin{equation}
\text{sim}(i, j) = D(j_y,f(i_x,i_y,j_x | \theta)) - D(j_y,f(j_x | \theta))
\label{delift_influence_function}
\end{equation}

\noindent After data valuation, the data selection stage consists of using submodular functions \citep{bilmes_submodularity}. In particular, we use the Facility Location submodular function. It takes as input a similarity kernel that will optimize the maximum similarity between the chosen subset and the overall dataset while also minimizing the size of the chosen subset. To minimize the subset size, the Facility Location -- and submodular functions, in general -- employ a diminishing gains property. This property states that samples added to a smaller subset have more value than samples added to a larger subset. Hence, we rely on our influence function to capture the informativeness of samples, and submodular functions to choose a set of representative samples, resulting in a small, information-rich subset on which to fine-tune a model.

\paragraph{DELIFT (SE)} \citep{delift} is a model-independent metric, and chooses samples from $\mathcal{D_F}$ which are semantically closest to the samples from $\mathcal{D_T}$. Semantic distance is calculated by the cosine distance between embeddings of samples:

\begin{equation}
\text{sim}(i, j) = \frac{<\text{emb}((i_x, i_y)), \text{emb}((j_x, j_y))>}{||\text{emb}((i_x, i_y))|| \cdot||\text{emb}((j_x, j_y))||}
\label{delift_se_influence_function}
\end{equation}

\noindent, where $\text{emb}(q)$ is an embedding model with input data $q$. Similar to DELIFT, DELIFT (SE) also uses the Facility Location function to select a small, information-rich subset of samples.

\paragraph{LESS} \citep{less} is model-dependent, gradient-based metric. Here, gradients between samples in $\mathcal{D_F}$ and $\mathcal{D_T}$ are matched by cosine similarity, and those that match the highest are chosen to represent $\mathcal{D_T}$ (where $\nabla(q; \theta)$ is the gradient of data point $q$ from a model with parameters $\theta$):

\begin{equation}
\text{sim}(i, j) = \frac{<\nabla((i_x, i_y); \theta), \nabla((j_x, j_y); \theta)>}{||\nabla((i_x, i_y); \theta)|| \cdot||\nabla((j_x, j_y); \theta)||}
\label{less_influence_function}
\end{equation}

\noindent During the data selection stage, the top-$k$ matching gradients are chosen to be part of the subset. One thing to notice is that the above equation implies a quadratic computation while Table \ref{table: costs} in the main text denotes a linear computation -- this is because the gradients for each data point only need to be computed once, while the cosine similarity can be computed many times inexpensively.

\subsection{Pointwise Influence Functions}
\label{app: pointwise_influence_functions}
Finally, \textbf{SelectIT} \cite{selectit} is another model-dependent metric that uses performance signals for data valuation, but incurs linear cost as it uses a model's uncertainty to rank data samples. Still, as mentioned in Table \ref{table: costs} from the main text, the linear time operations are forward propagations through LLMs.

SelectIT ranks data points based on their token-level, sentence-level, and model-level uncertainty expressed via token distribution. The token-level uncertainty is represented as the maximum probability of a token during next-token prediction. The sentence-level uncertainty is computed based on the token-level uncertainties of all the tokens in a sentence, for each prompt in a pool of prompts. Finally, the model-level uncertainty is calculated by taking a weighted average of the sentence-level uncertainty scores for multiple model sizes (the weights are determined by model size). This three-stage process provides a ranking process -- thus, during data selection, the points with the top-$k$ scores are chosen.

\section{License}
All the code of this project is under the Apache 2.0 License. The datasets MixInstruct and Alpaca are under the MIT and Creative Commons Attribution Non Commercial 4.0 International Licenses, respectively. The code for the baselines are under the MIT and Apache 2.0 Licenses. Our use of existing artifact(s) is consistent with their intended use. The artifacts are all in English, and do not contain data with personally identifiable information.

\newpage
\newpage
\section*{NeurIPS Paper Checklist}

\begin{enumerate}

\item {\bf Claims}
    \item[] Question: Do the main claims made in the abstract and introduction accurately reflect the paper's contributions and scope?
    \item[] Answer: \answerYes{} 
    \item[] Justification: All of our claims are supported by the empirical results throughout the paper.

\item {\bf Limitations}
    \item[] Question: Does the paper discuss the limitations of the work performed by the authors?
    \item[] Answer: \answerYes{} 
    \item[] Justification: We pose future work directions that address the limitations of our method in Section 6.

\item {\bf Theory assumptions and proofs}
    \item[] Question: For each theoretical result, does the paper provide the full set of assumptions and a complete (and correct) proof?
    \item[] Answer: \answerNA{} 
    \item[] Justification: Our paper does not make significant theoretical contributions to warrant proofs.

    \item {\bf Experimental result reproducibility}
    \item[] Question: Does the paper fully disclose all the information needed to reproduce the main experimental results of the paper to the extent that it affects the main claims and/or conclusions of the paper (regardless of whether the code and data are provided or not)?
    \item[] Answer: \answerYes{} 
    \item[] Justification: We not only include our code, but also detail the evaluation set up throughout Sections 4 and 5.

\item {\bf Open access to data and code}
    \item[] Question: Does the paper provide open access to the data and code, with sufficient instructions to faithfully reproduce the main experimental results, as described in supplemental material?
    \item[] Answer: \answerYes{} 
    \item[] Justification: We include an anonymized link to our code base in the abstract.

\item {\bf Experimental setting/details}
    \item[] Question: Does the paper specify all the training and test details (e.g., data splits, hyperparameters, how they were chosen, type of optimizer, etc.) necessary to understand the results?
    \item[] Answer: \answerYes{} 
    \item[] Justification: We detail our experimental settings throughout Sections 4 and 5.

\item {\bf Experiment statistical significance}
    \item[] Question: Does the paper report error bars suitably and correctly defined or other appropriate information about the statistical significance of the experiments?
    \item[] Answer: \answerYes{} 
    \item[] Justification: This analysis can be found in Table 2.

\item {\bf Experiments compute resources}
    \item[] Question: For each experiment, does the paper provide sufficient information on the computer resources (type of compute workers, memory, time of execution) needed to reproduce the experiments?
    \item[] Answer: \answerYes{} 
    \item[] Justification: We mention that we use 2 NVIDIA A40 GPUs.
    
\item {\bf Code of ethics}
    \item[] Question: Does the research conducted in the paper conform, in every respect, with the NeurIPS Code of Ethics \url{https://neurips.cc/public/EthicsGuidelines}?
    \item[] Answer: \answerYes{} 
    \item[] Justification: We followed the Code of Ethics.

\item {\bf Broader impacts}
    \item[] Question: Does the paper discuss both potential positive societal impacts and negative societal impacts of the work performed?
    \item[] Answer: \answerYes{} 
    \item[] Justification: We provide a summary of societal impacts in Section 6.
    
\item {\bf Safeguards}
    \item[] Question: Does the paper describe safeguards that have been put in place for responsible release of data or models that have a high risk for misuse (e.g., pretrained language models, image generators, or scraped datasets)?
    \item[] Answer: \answerNA{} 
    \item[] Justification: We only use open-source data and models under license (specified in Appendix D).

\item {\bf Licenses for existing assets}
    \item[] Question: Are the creators or original owners of assets (e.g., code, data, models), used in the paper, properly credited and are the license and terms of use explicitly mentioned and properly respected?
    \item[] Answer: \answerYes{} 
    \item[] Justification: Please refer to Appendix D.

\item {\bf New assets}
    \item[] Question: Are new assets introduced in the paper well documented and is the documentation provided alongside the assets?
    \item[] Answer: \answerYes{} 
    \item[] Justification: We release code, which is documented and under license.

\item {\bf Crowdsourcing and research with human subjects}
    \item[] Question: For crowdsourcing experiments and research with human subjects, does the paper include the full text of instructions given to participants and screenshots, if applicable, as well as details about compensation (if any)? 
    \item[] Answer: \answerNA{} 
    \item[] Justification: We do not use human subjects in this work.

\item {\bf Institutional review board (IRB) approvals or equivalent for research with human subjects}
    \item[] Question: Does the paper describe potential risks incurred by study participants, whether such risks were disclosed to the subjects, and whether Institutional Review Board (IRB) approvals (or an equivalent approval/review based on the requirements of your country or institution) were obtained?
    \item[] Answer: \answerNA{} 
    \item[] Justification: We do not use human subjects in this work.

\item {\bf Declaration of LLM usage}
    \item[] Question: Does the paper describe the usage of LLMs if it is an important, original, or non-standard component of the core methods in this research? Note that if the LLM is used only for writing, editing, or formatting purposes and does not impact the core methodology, scientific rigorousness, or originality of the research, declaration is not required.
    \item[] Answer: \answerYes{} 
    \item[] Justification: As this work is to improve LLMs, we clearly outline the (all open-source) LLMs that were used for experimentation purposes.

\end{enumerate}

\end{document}